\documentclass[letterpaper, 10 pt, conference]{ieeeconf}

\IEEEoverridecommandlockouts                              

\usepackage[nolist,nohyperlinks]{acronym}

\usepackage{xcolor}

\newcommand\submap[1]{\mathbb{#1}}

\usepackage{amsfonts}
\usepackage{amsmath}

\usepackage{graphicx}
\usepackage[font=footnotesize,labelfont=bf]{caption}
\usepackage{subcaption}

\usepackage{booktabs}
\newcommand{\ra}[1]{\renewcommand{\arraystretch}{#1}}

\usepackage{url}

\usepackage{mdframed}

\title{\LARGE \bf
Free-Space Features: Global Localization in 2D Laser SLAM Using Distance Function Maps
}
\author{Alexander Millane\textbf{}, Helen Oleynikova, Juan Nieto, Roland Siegwart, C\'{e}sar Cadena\\
 Autonomous Systems Lab, ETH  Z{\"u}rich
\thanks{This research was funded by the National Center of Competence in Research (NCCR) Robotics through the Swiss National Science Foundation.}\textbf{}
}

\begin{acronym}
\acro{SLAM}{Simultaneous Localization And Mapping}
\acro{SDF}{Signed Distance Function}
\acro{DoH}{Determinant of Hessian}
\acro{BoW}{Bag of Words}
\end{acronym}

\begin{document}
\maketitle

\begin{abstract}
In many applications, maintaining a consistent map of the environment is key to enabling robotic platforms to perform higher-level decision making. Detection of already visited locations is one of the primary ways in which map consistency is maintained, especially in situations where external positioning systems are unavailable or unreliable. Mapping in 2D is an important field in robotics, largely due to the fact that man-made environments such as warehouses and homes, where robots are expected to play an increasing role, can often be approximated as planar. Place recognition in this context remains challenging: 2D lidar scans contain scant information with which to characterize, and therefore recognize, a location. This paper introduces a novel approach aimed at addressing this problem. At its core, the system relies on the use of the distance function for representation of geometry. This representation allows extraction of features which describe the geometry of both surfaces \emph{and} free-space in the environment. We propose a feature for this purpose. Through evaluations on public datasets, we demonstrate the utility of free-space in the description of places, and show an increase in localization performance over a state-of-the-art descriptor extracted from surface geometry.
\end{abstract}

\section{Introduction}
\label{sec:introduction}

A key competency towards achieving high-level tasks is the ability for a robot to build an internal representation of the environment and localize within it. Systems performing this task, known as \ac{SLAM}, are increasingly being deployed on robots operating in unstructured environments or without access to reliable external localization infrastructure~\cite{cadena2016past}.

Ground-robots frequently operate in environments which can be approximated as locally-planar and therefore 2D \ac{SLAM} is an important field in robotic research. Global-localization in this context, however, remains challenging. Some effort has been made to design local descriptors for 2D lidar data, drawing inspiration from the techniques that have made place recognition successful in visual \ac{SLAM}. The primary challenge, however, is that 2D scans, in contrast to images, contain scarce information about the environment, complicating efforts to design sufficiently powerful descriptors with which to characterize \textit{places}.

Existing local descriptors for 2D lidar data are typically constructed from collections of points on the surface of occupied space. As we will show, this formulation, however, omits available information from the description of place. One of the primary purposes of mapping systems is to determine regions of free-space, such that navigation can be conducted in these areas. The hypothesis motivating this work is that these regions might also be meaningful for localization; that is, the shape and arrangement of the free-space is likely to contain substantial information about a place.



\begin{figure}[t] 
    \centering
    \begin{subfigure}[b]{1.00\columnwidth}
        \includegraphics[trim={0cm 0cm 0cm 0cm},clip,width=\columnwidth]{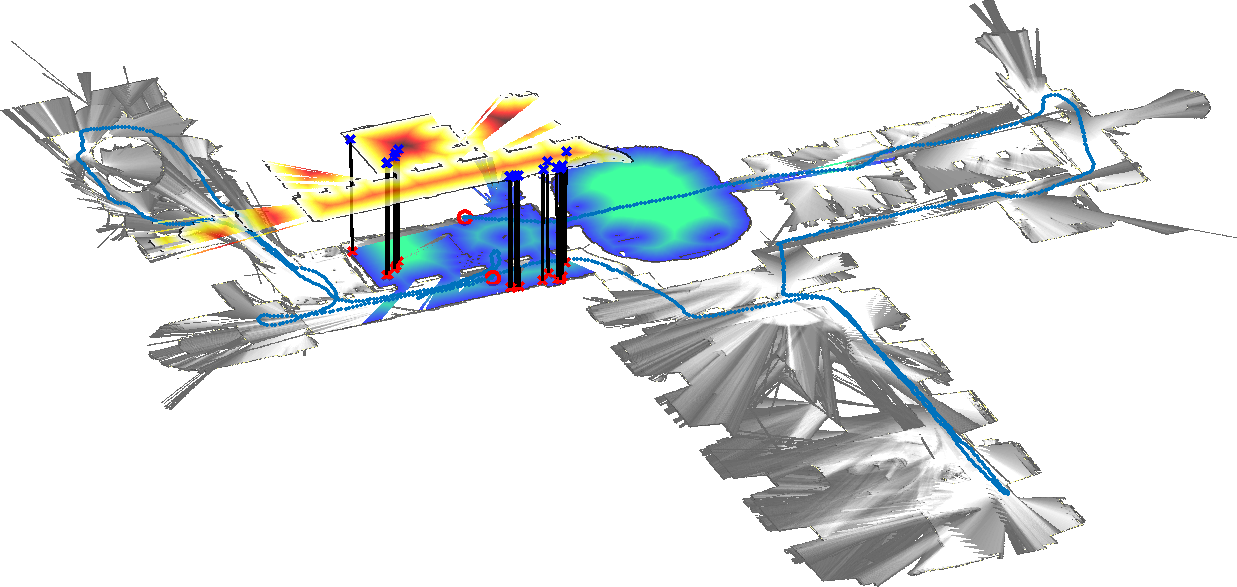}
        \caption{}
        \label{fig:teaser_main}
    \end{subfigure}
    \begin{subfigure}[b]{1.00\columnwidth}
        \includegraphics[trim={0cm 0cm 6.5cm 0cm},clip,width=\columnwidth]{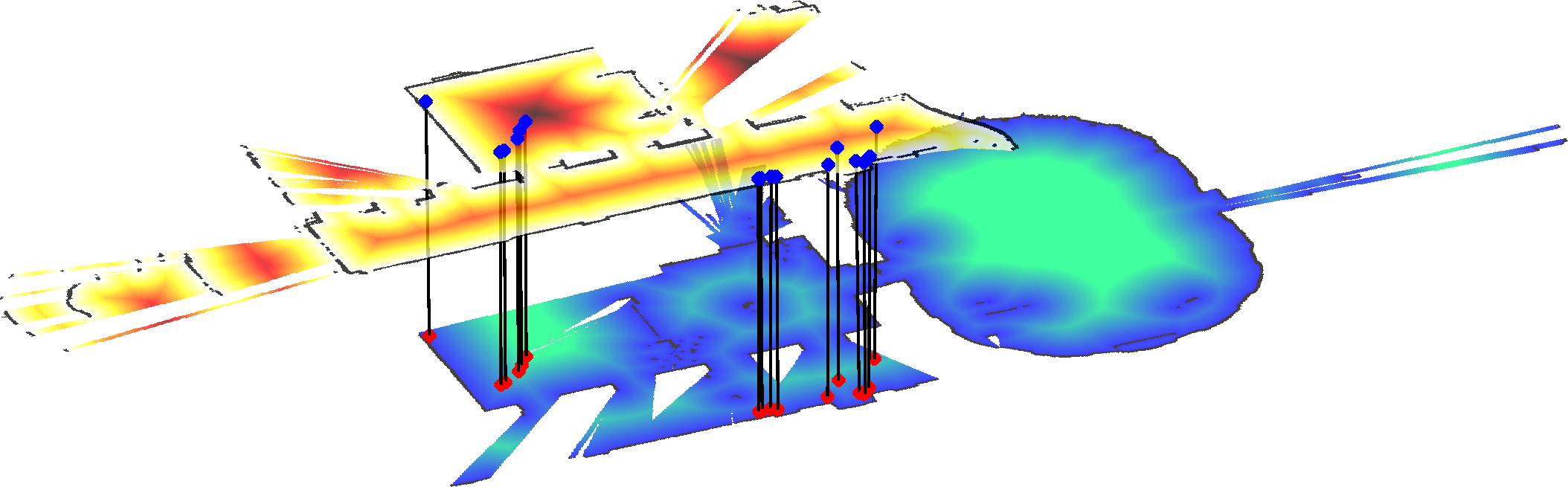}
        \caption{}
        \label{fig:teaser_close_up}
    \end{subfigure}
    \caption{An example of place-recognition using the proposed method. The query submap (red) from the end of a trajectory is matched against a submap (blue) from the start. Query and match submaps are displayed as distance functions, free-space feature matches are shown in black, and the path taken by the agent in blue. A close up of the matched submaps is shown in  (\subref{fig:teaser_close_up}).}
    \label{fig:teaser}
    \vspace{-5mm}
\end{figure}

In this paper we introduce a novel feature aimed at this purpose. Central to our system is the representation of the geometry of the mapped world using a \ac{SDF}. Distance functions have been used in a wide variety of applications, including robot path planning~\cite{lau2013efficient}, dense reconstruction~\cite{izadi2011kinectfusion}, and computer graphics~\cite{frisken2000adaptively}. In contrast to pointclouds, the distance function represents the geometry of free and occupied space equally. We posit that by extracting keypoints and descriptors on the \ac{SDF}, we characterize the local geometry of both free and occupied space. We suggest a simple keypoint detector and descriptor for that purpose and test the efficacy of the resulting approach. In summary, the contributions of this paper are:
\begin{itemize}
    \item The use of distance functions explicitly for the purpose of place recognition.
    \item The development of a keypoint detection and description approach for characterization of local \ac{SDF} geometry.
    \item Experimental validation on publicly available datasets showing the efficacy of the proposed feature for global localization, as well as an analysis of the contribution of free-space to its performance.
\end{itemize}

\section{Related Work}
\label{sec:related_work}

Several works have addressed the problem of localization in 2D lidar maps. Various approaches have been taken for which we give a brief overview.

\subsection{Scan Matching}

One common approach is to match incoming lidar data to an existing map using scan-matching techniques, similar to those used to incrementally track robot motion~\cite{grisetti2007improved}. The approach of scan-matching against \emph{many} past scans, however, becomes computationally burdensome, and several recent works have suggested techniques to improve its scalability. Google's Cartographer~\cite{hess2016real} uses a branch-and-bound method for eliminating bad scan-to-map matches early, leading to an efficient implementation which is widely used. Similarly, Olson~\cite{olson2015m3rsm} uses a multi-resolution technique to speed up an exhaustive search. Several works suggest the use of correlation-based techniques for matching recent scans to an existing map~\cite{gutmann1999incremental, olson2009real}. The principal weakness of these techniques, however, is that the cost to search the whole map becomes prohibitive as the map grows, even to a moderate size. In practice, therefore, these systems restrict the search space to an area surrounding the current pose, limiting the scope of applicability for \emph{global}-localization.

\subsection{Scan Description}

Drawing inspiration from image-retrieval literature~\cite{sivic2003video}, several works have proposed local descriptors for 2D lidar data~\cite{walthelm2002new,tipaldi2010flirt,tipaldi2013geometrical}. Places are characterized as collections of descriptors which can then be stored in a database and efficiently searched. Tipaldi et al.~\cite{tipaldi2010flirt} introduced FLIRT features and demonstrated the efficacy of the image-retrieval style approach. In similarity to this work, regions within FLIRT feature support observed as free contribute to the description, however, the descriptors are fundamentally designed to describe the geometry of the surface on which they are extracted. In keeping with the image-retrieval metaphor, the authors extended their work~\cite{tipaldi2013geometrical} to follow a \ac{BoW}~\cite{sivic2003video} style approach, performing clustering in descriptor space, replacing raw features with cluster membership, and describing places as a combination of these \textit{words}. The resulting dimensionality reduction reduces the search complexity and allows extension to larger scale environments~\cite{tipaldi2013geometrical}. Recently, learning has also been applied to the problem of matching pairs of lidar scans~\cite{granstrom2011learning, li2017deep}. We draw inspiration from the image retrieval style approach of these works and aim to extend descriptions to explicitly include free-space geometry.

\subsection{Submap Description}

Submapping has long been applied to improve the scalability and consistency of maps produced by \ac{SLAM} systems~\cite{bosse2004simultaneous, millane2018c}. In localization as well, authors have suggested the use of sub-mapping approaches to reduce ambiguities in place-recognition using single scans alone~\cite{gutmann1999incremental}. Bosse and Zlot~\cite{bosse2009keypoint} pose the place recognition problem as one of recognizing a revisited \emph{submap} rather than a revisited scan. The authors of this work compare several keypoint and description methods and demonstrate the efficacy of the approach to large-scale datasets including trajectories of 10s of kilometeres and $\approx$1000 submaps. We draw inspiration from this work and extend it with a new feature type.

\subsection{Signed Distance Functions}

Lastly, \ac{SDF}s are widey used in the domain of path planning~\cite{lau2013efficient, oleynikova2017voxblox} where the representation allows efficient collision checking and topology extraction. For mapping applications, \ac{SDF}s have experienced a resurgence in recent years where they have proven useful for aggregating visual data from consumer-grade depth cameras~\cite{izadi2011kinectfusion}. Following this development, several recent works~\cite{fossel20152d,koch2016multi} have investigated the utility of \ac{SDF}s in the front end of 2D lidar mapping systems. This paper is an investigation into the utility of this representation for place-recognition in that context.

\section{Problem Statement}
\label{sec:problem_statement}
Given a sequence of scans from a 2D laser rangefinder, we use an existing approach~\cite{hess2016real} to produce a sequence of locally-consistent submaps $\{\submap{S}_k\}_{k=1}^N$ and associated coordinate frames $\{S_k\}_{k=1}^N$. We aim to determine if two submaps $\submap{S}_i$ and $\submap{S}_j$ correspond to the same location, that is, contain significant overlap in the region of the environment they describe, and to determine their relative transformation $T_{S_i S_j} \in \mathrm{SE}(2)$. Central to our approach is the use of the \ac{SDF}, which we denote as the function $f: \mathbb{R}^2 \to d$, where $d \in \mathbb{R}$ is the signed distance to the nearest surface. As is common in recent reconstruction systems, we store $f$ as a collection of samples over a discrete uniformly-spaced voxel grid.

\section{Approach}
\label{sec:approach}

In this section we describe our approach for submap representation, keypoint detection and description.

\subsection{Submap representation}

\begin{figure}[t] 
    \centering
    \begin{subfigure}[b]{0.45\columnwidth}
        \includegraphics[trim={0cm 0cm 0cm 0cm},clip,width=\columnwidth]{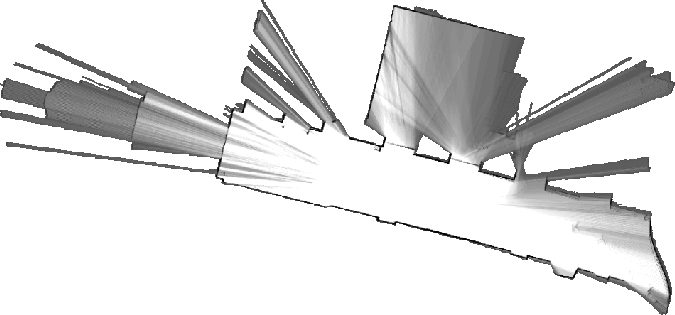}
        \caption{}
        \label{fig:approach_prob_grid}
    \end{subfigure}
    \begin{subfigure}[b]{0.45\columnwidth}
        \includegraphics[trim={0cm 0cm 0cm 0cm},clip,width=\columnwidth]{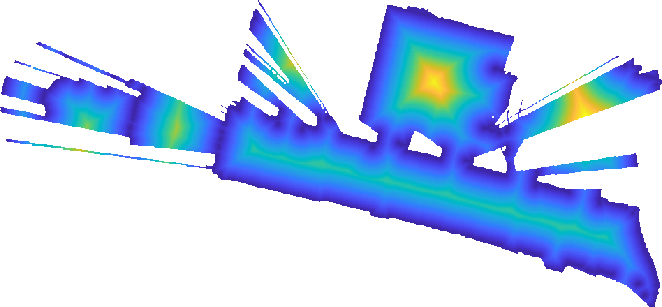}
        \caption{}
        \label{fig:approach_sdf_grid}
    \end{subfigure}
    \begin{subfigure}[t]{0.075\columnwidth}
        \includegraphics[trim={0cm 0cm 0cm 0cm},clip,width=\columnwidth]{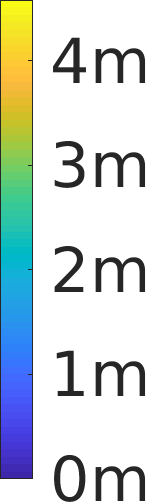}
    \end{subfigure}
    \caption{An example submap, taken from the Deutsches Museum datasets~\cite{hess2016real}, represented as a occupancy probability grid (\subref{fig:approach_prob_grid}), and an \ac{SDF} (\subref{fig:approach_sdf_grid}).}
    \label{fig:prob_grid_to_sdf}
    \vspace{-5mm}
\end{figure}

Input submaps from the \ac{SLAM} front-end~\cite{hess2016real} are initially parameterized as occupancy probability grids, a function mapping from observed space (discretized into voxels), $\Omega \subset \mathbb{Z}^2$, to a probability of occupancy, and unknown space to an sentinel value. We generate an \ac{SDF} by thresholding the probability to produce a binary-valued grid, and then by taking the distance transform using the algorithm described in~\cite{maurer2003linear}. Figure~\ref{fig:prob_grid_to_sdf} shows the results of this process for an example submap.

\subsection{Keypoint Detection}
\label{sec:approach_selection}

\begin{figure}[t] 
    \centering
    \includegraphics[trim={0cm 0cm 0cm 0cm},clip,width=0.75\columnwidth]{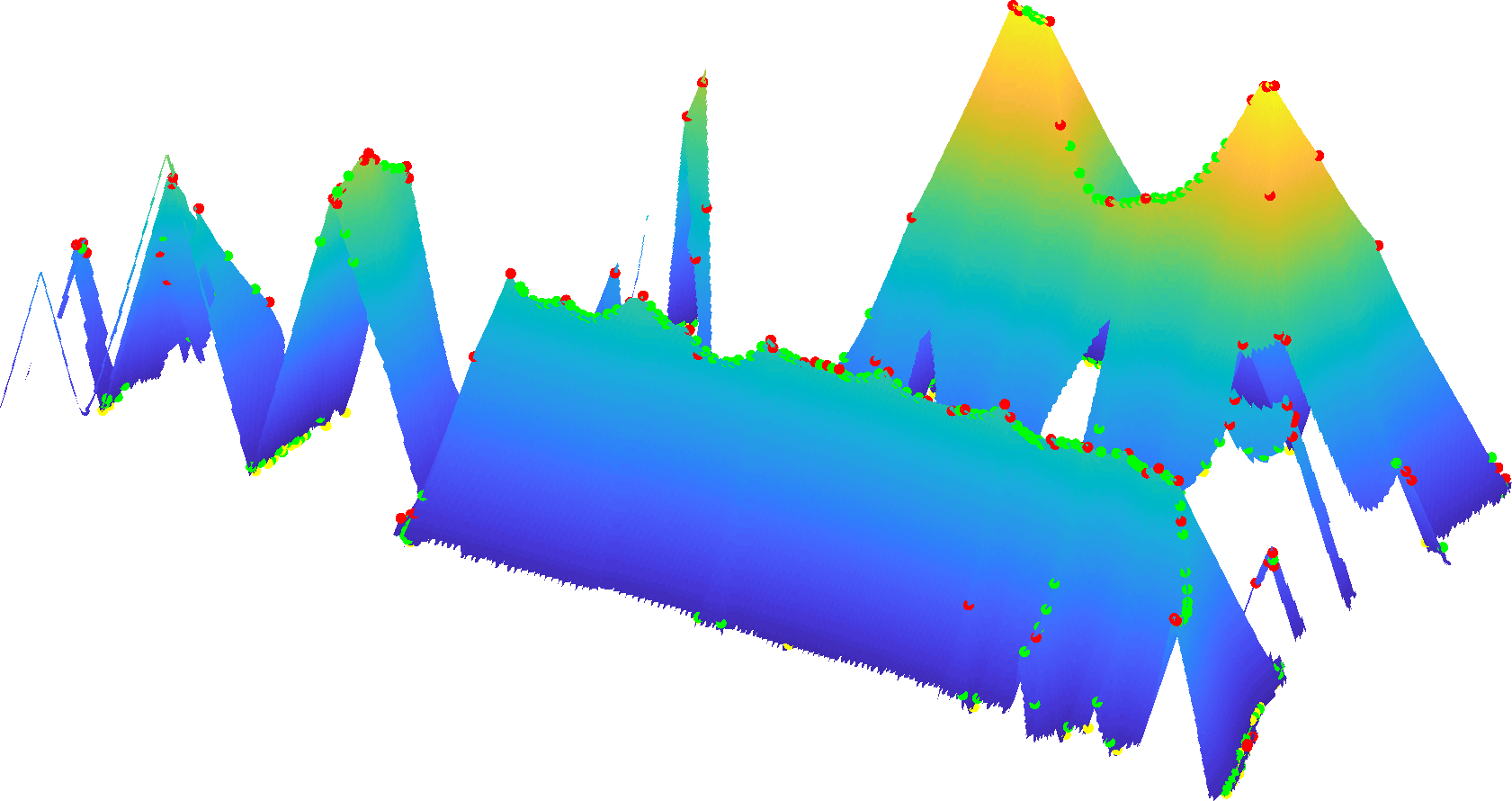}
    \caption{The \ac{SDF}, $f: \mathbb{R}^2 \to d$ with $d\in\mathbb{R}$, of the example submap from Fig.~\ref{fig:prob_grid_to_sdf} represented as surface in $\mathbb{R}^3$. Also shown are the extracted keypoints, classified as local maximums (red), saddles (green) and minimums (yellow).}
    \label{fig:keypoint_detection_surface}
\end{figure}

Keypoint detection aims to extract points which are salient, that is can be reliably re-extracted, and interesting enough to warrant description. In contrast to image data, an \ac{SDF} is by definition smooth, in the sense that it is differentiable \textit{almost everywhere}. The result is that the \ac{SDF} is free from abrupt changes in value, which are the typical candidates for keypoints in images. We therefore select points of high curvature using a detector based on the Hessian of the \ac{SDF}. In particular, we use the \ac{DoH}, an approximation of which is used to detect blobs in the popular SURF descriptor for images~\cite{bay2006surf}. We compute the Hessian of the distance function
\begin{equation}
    H = 
    \begin{bmatrix}
        H_{xx} & H_{xy} \\
        H_{yx} & H_{yy} \\
    \end{bmatrix}
\end{equation}
where,
\begin{align}
    H_{xx} &= \frac{\partial^2}{\partial x^2} (f * G),  \nonumber \\
    H_{yy} &= \frac{\partial^2}{\partial y^2} (f * G), \\ 
    H_{xy} &= \frac{\partial^2}{\partial x \partial y} (f * G),  \nonumber
\end{align}
where G is a Gaussian kernel with a tunable variance $\sigma^2$. This operation is performed by convolving the distance data with the Sobel derivative kernel. The \ac{DoH} is extracted as
\begin{equation}
    det(H) = H_{xx} H_{yy} - H_{xy}^2.
\end{equation}
We then perform a search which selects areas of locally maximal Gaussian curvature on the \ac{SDF} (see Fig.~\ref{fig:keypoint_detection_surface}). Note that care is taken to not detect features on the barrier between observed and unobserved space. At this stage we also classify points passing the selection process based on the surface topology on which they're extracted. In particular, we calculate the eigenvalues
\begin{equation}
\lambda_1, \lambda_2 = \text{eig}(H),
\end{equation}
and classify points as either maxima, minima or saddles based on their signs. This classification becomes part of their description (Sec.~\ref{sec:approach_decription}).

\subsection{Keypoint Description}
\label{sec:approach_decription}
\begin{figure}[t] 
    \centering
    \begin{subfigure}[b]{0.5\columnwidth}
        \includegraphics[trim={0cm 0cm 0cm 0cm},clip,width=\columnwidth]{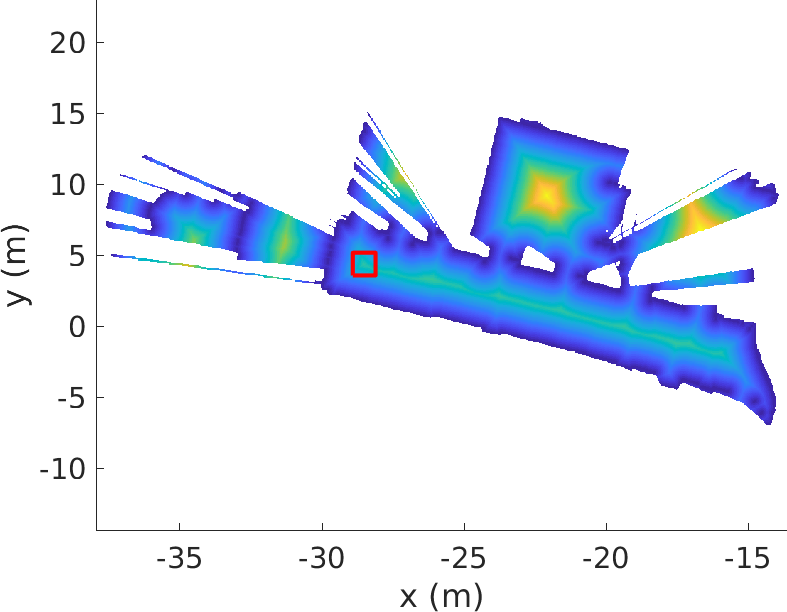}
        \caption{}
        \label{fig:feature_location}
    \end{subfigure}
    \begin{subfigure}[b]{0.5\columnwidth}
        \includegraphics[trim={0cm 0cm 0cm 0cm},clip,width=\columnwidth]{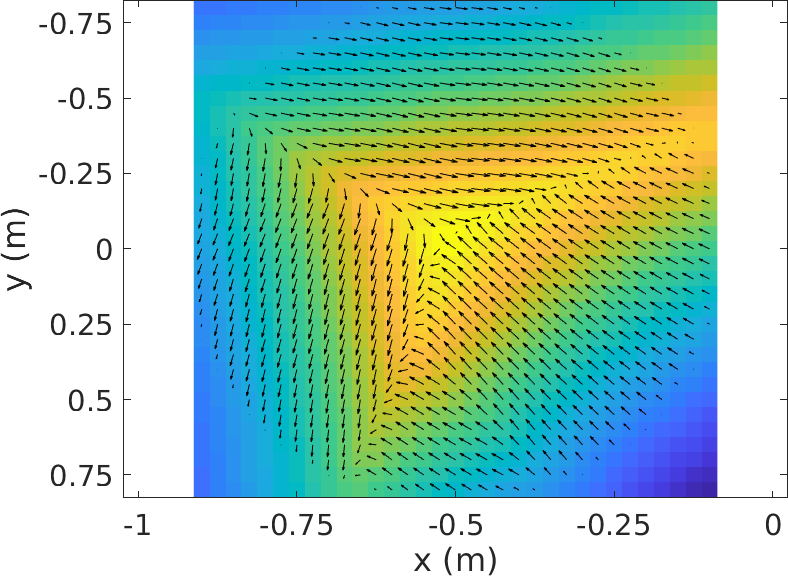}
        \caption{}
        \label{fig:feature_patch}
    \end{subfigure}
    \begin{subfigure}[b]{0.5\columnwidth}
        \includegraphics[trim={0cm 0cm 0cm 0cm},clip,width=\columnwidth]{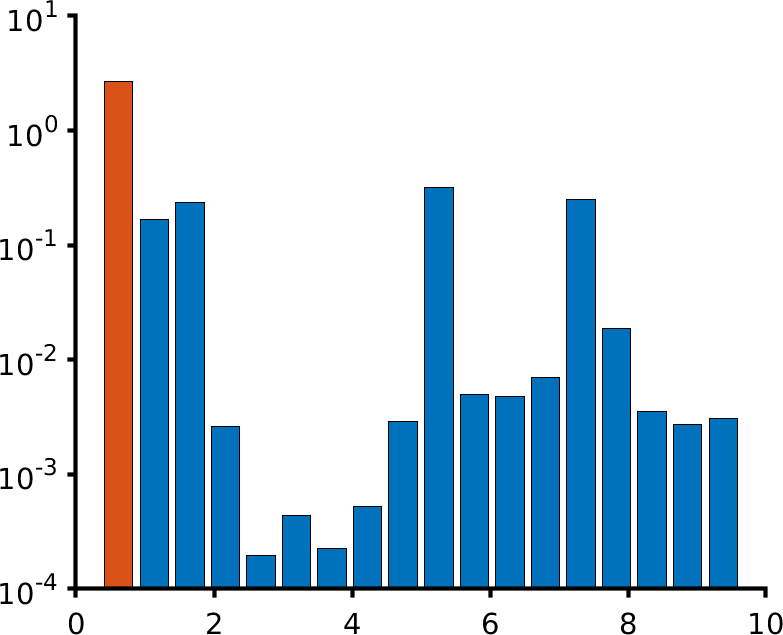}
        \caption{}
        \label{fig:feature_histogram}
    \end{subfigure}
    \caption{An example feature extracted from an \ac{SDF} submap showing (\subref{fig:feature_location}) the location of the detected feature, (\subref{fig:feature_patch}) weighted gradients in a circular window, and (\subref{fig:feature_histogram}), the computed orientation histogram (blue) and the average \ac{SDF} value of the extracted window (red).}
    \label{fig:feature}
    \vspace{-5mm}
\end{figure}

Taking inspiration from SIFT~\cite{dalal2005histograms} and HOG~\cite{lowe1999object} image features, we construct a descriptor based partially on a histogram of gradient orientations. We first extract a circular window of \ac{SDF} values around a selected keypoint and compute gradient orientations and magnitudes in this window, shown in Fig.~\ref{fig:feature_location} and Fig.~\ref{fig:feature_patch}. To achieve rotational invariance we use a 36-bin gradient orientation histogram to determine the dominant orientation within this window and express orientations relative to this direction, the approach used by SIFT~\cite{lowe1999object}. We then construct a 17-bin histogram using the relative gradient orientations (see Fig. \ref{fig:feature_histogram}), which forms the first part of our descriptor. The gradients contributing to these histograms are weighted by their magnitude and a Gaussian kernel which provides higher weights to central gradients.

In addition to the orientation histogram, we attach a weighted measurement of the window's average \ac{SDF} value. In practice, the inclusion of the distance means that during lookup features with substantially different average distance function values will not be retrieved, which we found to increase descriptor performance. Note that in contrast to image features, where pixels values within the descriptor support are subject to substantial changes between observations (for example, due to lighting changes), the distance produced by $f$ are metrically scaled. To further restrict matches, we require matched features to have the same classification (see Sec.~\ref{sec:approach_selection}), as these keypoints represent areas in the environment with distinct topology. Maximums are extracted in areas between obstacles, minimums on surface boundaries and saddles on geometric restrictions (areas where distance from obstacles grows in one direction but reduces in the other).

\subsection{Place recognition}

In this work we consider place recognition as the pairwise matching problem - that is, given two submaps $\submap{S}_i$ and $\submap{S}_j$ determine if they are matching. For each submap pair we determine feature correspondences using a nearest neighbour lookup, rejecting ambiguous matches using the \textit{ratio-test}~\cite{lowe1999object}. This test discards matches which have a nearby second neighbour. We use RANSAC to determine inlier correspondences as well as a $\mathrm{SE}(2)$ transform relating the submaps. If the number of inliers exceeds a threshold, the pair are considered a match. Note that, in application, place-recognition systems use several techniques to avoid pairwise comparisons and speed up lookups, such as inverted files and descriptor clustering~\cite{sivic2003video}, as well as approximate nearest neighbour voting~\cite{bosse2009keypoint}. Similarly, the authors in~\cite{bosse2009keypoint} suggest several match verification steps, which in practice are advisable given the potentially disastrous consequences of false matches for map consistency. These techniques are equally applicable here, however, their evaluation is beyond the scope of this paper. 
\section{Results}
\label{sec:results}

The results presented in this section aim to validate the hypotheses of this paper, that a) features extracted from \ac{SDF} submaps can be used for place recognition, and b) features in free-space improve localization performance when compared to features using only occupied space.

\begin{table}[t]
\centering
\ra{1.3}
\begin{tabular}{@{}rrrrr@{}}
\toprule
\multicolumn{2}{c}{Ours} && \multicolumn{2}{c}{Shape Contexts~\cite{belongie2002shape}} \\
\cmidrule{1-2} \cmidrule{4-5}
Parameter & value && Parameter & value \\
\cmidrule{1-2} \cmidrule{4-5}
radius & 0.8\,m && radius & 2.0\,m\\
number bins & 17 && number radial bins & 3 \\
& && number angular bins & 6 \\
distance weight & 0.002 \\
detection threshold & 0.0025 && detection threshold & 0.05 \\
matching max ratio & 0.75 && matching max ratio & 0.75 \\
\bottomrule
\end{tabular}
\caption{Major parameters for the proposed and comparison methods used to generate the results is Sec.\ref{sec:results}. Parameter settings for both methods were determined using a grid search for optimal performance on a localization experiment.}
\label{tab:tunings}
\vspace{-5mm}
\end{table}

\begin{table}[t]
\centering
\ra{1.3}
\begin{tabular}{@{}rrrr@{}}
\toprule
& & \multicolumn{2}{c}{Recall at Precision 1.0} \\
\cmidrule{3-4} 
Trajectory & \# Submaps & Ours & Shape Contexts~\cite{belongie2002shape} \\
\midrule
EG & 53 & \textbf{0.45} & 0.27 \\
OG & 90 & \textbf{0.18} & 0.07 \\
UG & 84 & \textbf{0.30} & 0.18 \\
PR1 & 36 & \textbf{0.85} & 0.33 \\
PR2 & 54 & \textbf{0.76} & 0.29 \\
PR3 & 71 & \textbf{0.99} & 0.36 \\
\bottomrule
\end{tabular}
\caption{Results for localization experiments discussed in Sec.~\ref{sec:results_descriptor_performance}. The tabulated numbers indicate the maximum achieved recall over inlier thresholds achieving precision 1.0.}
\label{tab:results_museum}
\vspace{-5mm}
\end{table}

\subsection{Descriptor Performance}
\label{sec:results_descriptor_performance}

To quantify the performance of the proposed method, we perform evaluations on the \textit{Deutsches Museum} dataset\footnote{\url{https://google-cartographer-ros.readthedocs.io/en/latest/data.html}}, available as part of Google's Cartographer \ac{SLAM} system~\cite{hess2016real}, as well as the PR2 Willow Garage dataset~\cite{mason2012object}. The former was captured by a backpack mounted lidar system as a person walked around a museum, and the later was captured by a mobile robot navigating an office environment. We use the cartographer system to generate a globally optimized and consistent map, and take the submap poses as ground-truth for our evaluation. Note that in these datasets no wide-baseline loop closures are performed; Cartographer maintains map consistency by finding scan-to-map loop closures (see Sec.~\ref{sec:related_work}). In contrast, we solve the more challenging \textit{global}-localization problem, and all localizations are performed without pose priors.

\begin{figure*}[t] 
    \centering
    \begin{subfigure}[b]{0.22\textwidth}
        \includegraphics[trim={0cm 0cm 0cm 0cm},clip,width=\columnwidth]{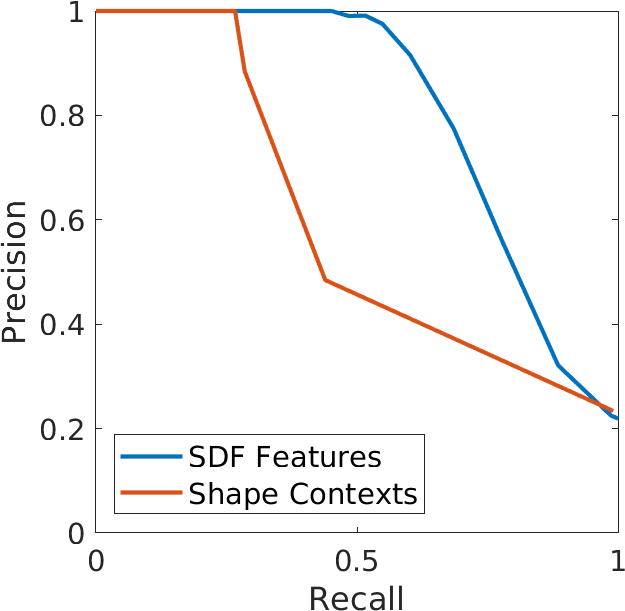}
    \end{subfigure}
    \begin{subfigure}[b]{0.22\textwidth}
        \includegraphics[trim={0cm 0cm 0cm 0cm},clip,width=\columnwidth]{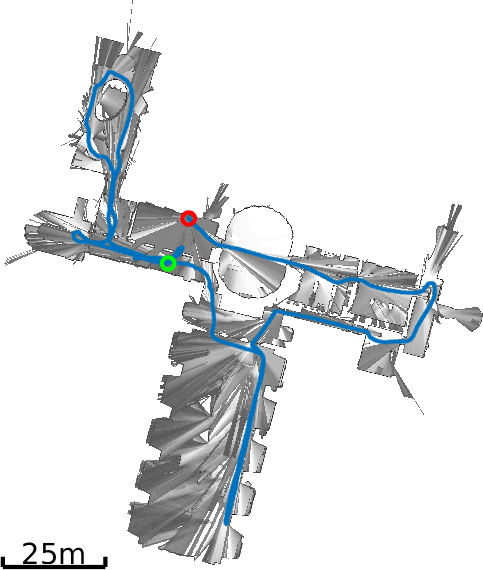}
        \caption{}
        \label{fig:carto_1}    
    \end{subfigure}
    \begin{subfigure}[b]{0.22\textwidth}
        \includegraphics[trim={0cm 0cm 0cm 0cm},clip,width=\columnwidth]{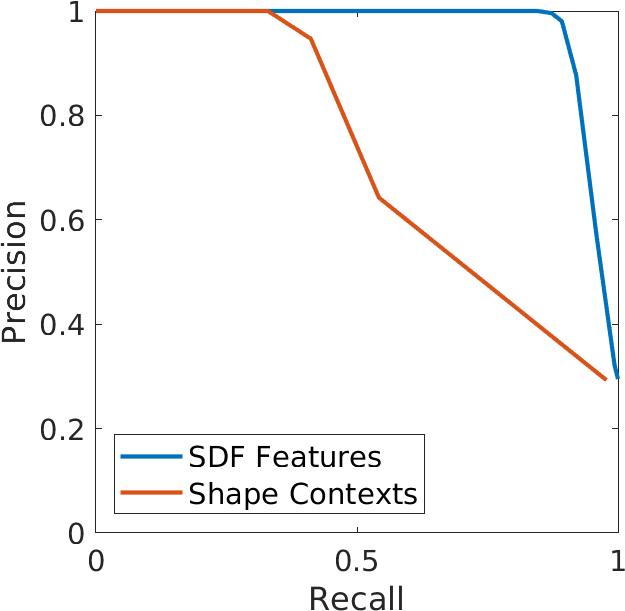}
    \end{subfigure}
    \begin{subfigure}[b]{0.22\textwidth}
        \includegraphics[trim={0cm 0cm 0cm 0cm},clip,width=\columnwidth]{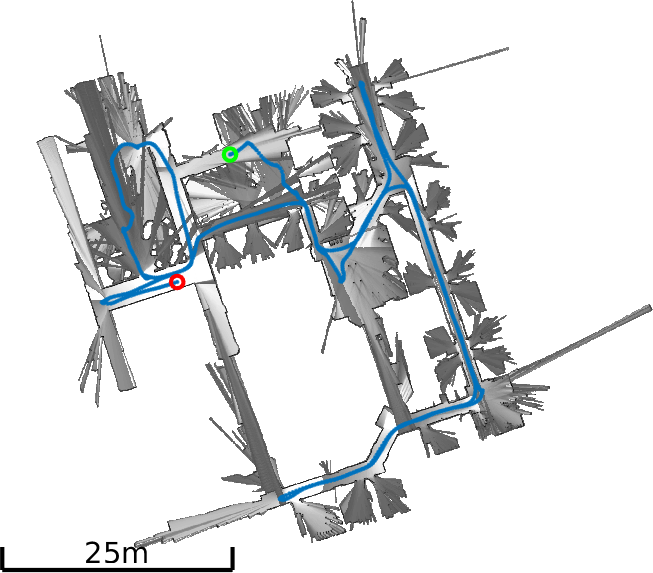}
        \caption{}
        \label{fig:pr_1}
    \end{subfigure}
    \begin{subfigure}[b]{0.22\textwidth}
        \includegraphics[trim={0cm 0cm 0cm 0cm},clip,width=\columnwidth]{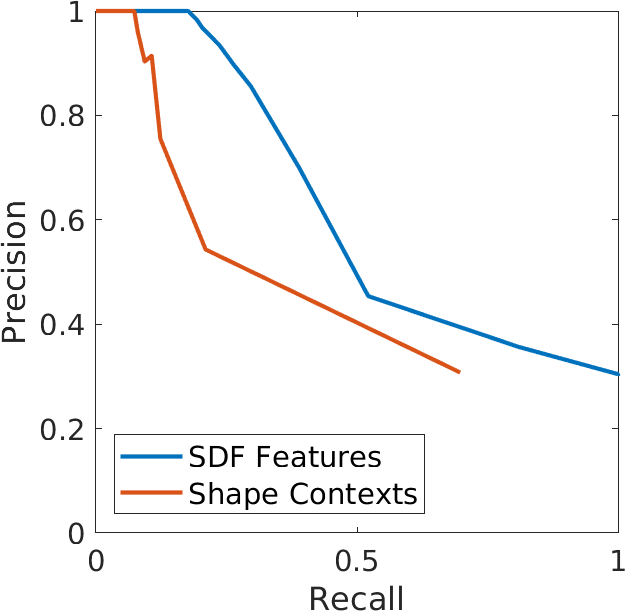}
    \end{subfigure}
    \begin{subfigure}[b]{0.22\textwidth}
        \includegraphics[trim={0cm 0cm 0cm 0cm},clip,width=\columnwidth]{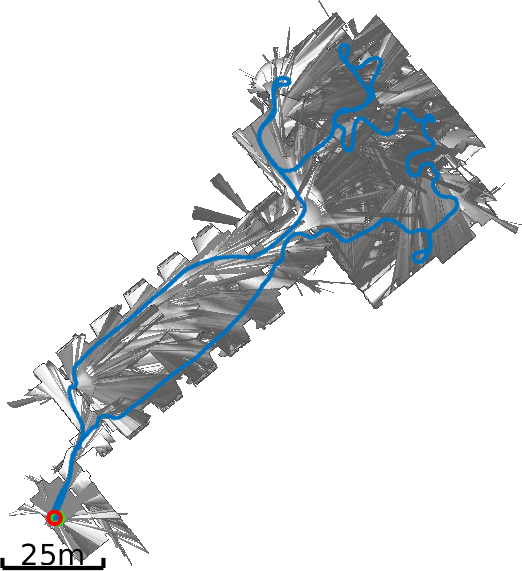}
        \caption{}
        \label{fig:carto_2}
    \end{subfigure}
    \begin{subfigure}[b]{0.22\textwidth}
        \includegraphics[trim={0cm 0cm 0cm 0cm},clip,width=\columnwidth]{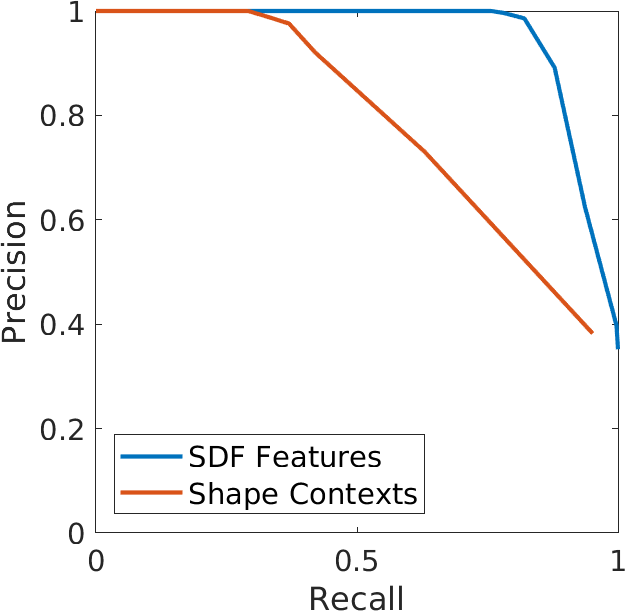}
    \end{subfigure}
    \begin{subfigure}[b]{0.22\textwidth}
        \includegraphics[trim={0cm 0cm 0cm 0cm},clip,width=\columnwidth]{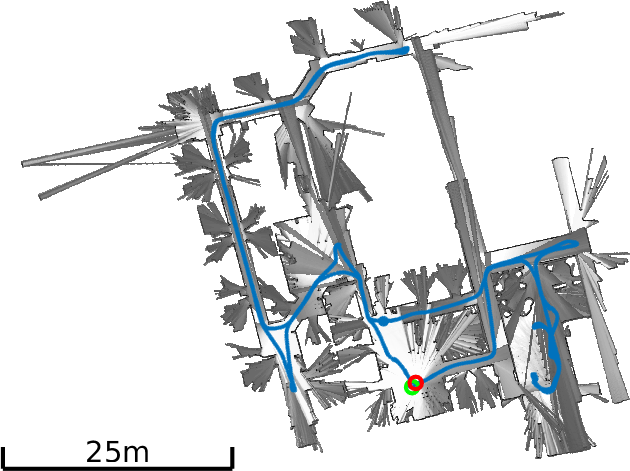}
        \caption{}
        \label{fig:pr_2}
    \end{subfigure}
    \begin{subfigure}[b]{0.22\textwidth}
        \includegraphics[trim={0cm 0cm 0cm 0cm},clip,width=\columnwidth]{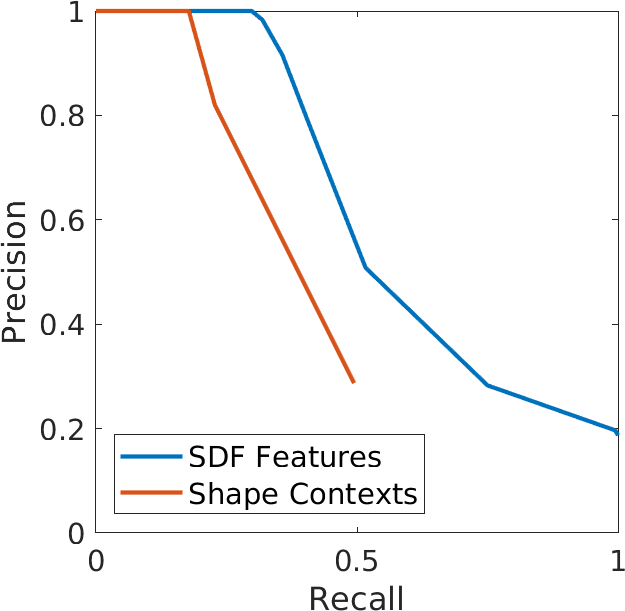}
    \end{subfigure}
    \begin{subfigure}[b]{0.22\textwidth}
        \includegraphics[trim={0cm 0cm 0cm 0cm},clip,width=\columnwidth]{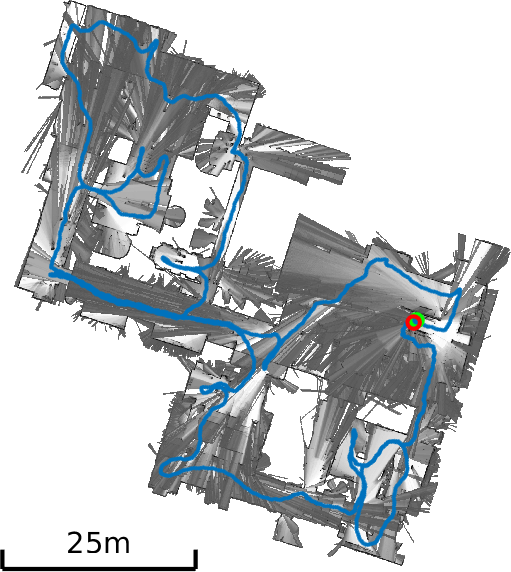}
        \caption{}
        \label{fig:carto_3}
    \end{subfigure}
    \begin{subfigure}[b]{0.22\textwidth}
        \includegraphics[trim={0cm 0cm 0cm 0cm},clip,width=\columnwidth]{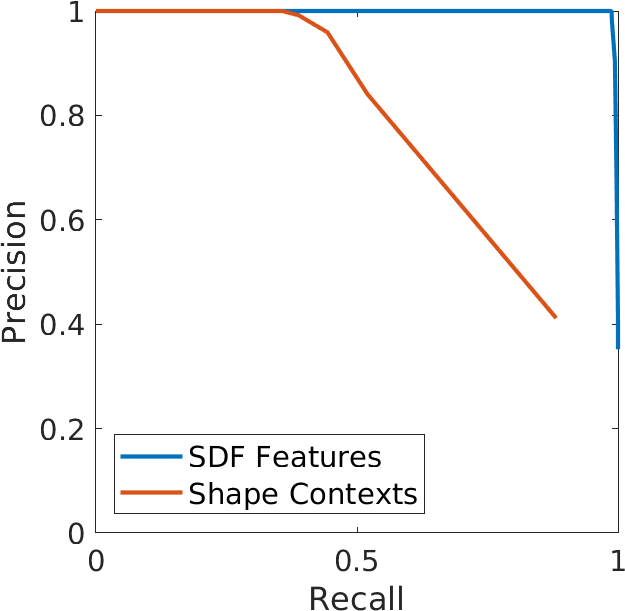}
    \end{subfigure}
    \begin{subfigure}[b]{0.22\textwidth}
        \includegraphics[trim={0cm 0cm 0cm 0cm},clip,width=\columnwidth]{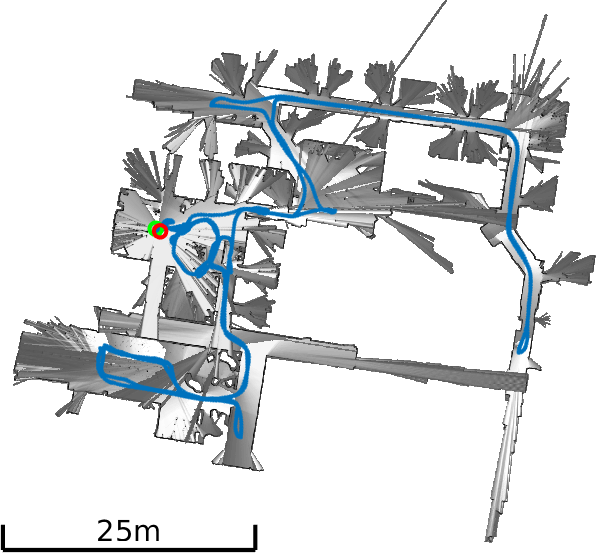}
        \caption{}
        \label{fig:pr_3}
    \end{subfigure}
    \caption{Precision-Recall curves for the proposed method (blue) and a comparison method (red) (a combination of curvature-clusters and Shape Contexts~\cite{belongie2002shape}). Maps are created using 2D lidar data from the Deutsches Museum dataset~\cite{hess2016real} (\subref{fig:carto_1}), (\subref{fig:carto_2}), (\subref{fig:carto_3}) and the Willow Garage PR2 dataset~\cite{mason2012object} (\subref{fig:pr_1}), (\subref{fig:pr_2}), (\subref{fig:pr_3}).}
    \label{fig:precision_recall_sdf_vs_sc}
    \vspace{-5mm}
\end{figure*}

We generate precision-recall curves in the following manner. First, we randomly select 1000 non-duplicate submap pairs and rotate them such that the pair has a random relative orientation. Ground truth match/non-match labels are determined using the ground-truth submap overlap; pairs with a sufficient proportion of overlapping observed voxels are considered matching. We perform submap matching and alignment using the proposed method and vary the inlier threshold to produce precision-recall curves.

In addition to our method, we test a state-of-the-art detector and descriptor combination that showed very good performance in an evaluation of many such combinations~\cite{bosse2009keypoint}. In particular, we implemented \text{Curvature Clusters} for keypoint selection and \textit{Shape Contexts}~\cite{belongie2002shape} for their description. These features require a pointcloud rather than a gridded representation. To produce input data, we take a globally optimized trajectory and aggregate back-projected undistorted scans to produce a pointcloud corresponding to each submap. We perform temporal sub-sampling on the scans such that each submap contains a reasonable number of points ($\approx$30 scans/submap). Note that the globally optimized trajectory is only used for the pointcloud generation, and it is not used for \ac{SDF} submap generation. 

Prior to analysis we performed a grid search over major algorithm parameters to determine the settings which resulted in the best performance for both the proposed and comparison method. Parameters were selected to maximize the recall score for precision 1.0 on a separate localization experiment (see Table~\ref{tab:tunings} for parameter values).

Figure~\ref{fig:precision_recall_sdf_vs_sc} shows the results of this analysis for three trajectories in the Deutsches Museum and three trajectories from the PR2 dataset (denoted \textit{EG}, \textit{OG}, \textit{UG}, \textit{PR1}, \textit{PR2}, and \textit{PR3} for brevity\footnote{Datasets referenced: OG: b2-2014-12-03-10-40-04, EG: b2-2016-04-05-14-44-52, UG: b2-2015-08-18-11-55-04, PR1: 2011-09-15-08-32-46, PR2: 2011-09-12-14-47-01, PR3: 2011-08-31-20-44-19}). The \ac{SLAM} front-end generates 53, 90, 84, 36, 54, and 71 submaps on these trajectories respectively. All trajectories revisit previously mapped areas frequently, and the datasets contain substantial opportunity for place-recognition. Table~\ref{tab:results_museum} shows the maximum recall over inlier match thresholds which generate precision 1.0. The proposed method achieve recall values of 0.45, 0.18, 0.30, 0.85, 0.76, and 0.99 for each of the datasets, an increase of 69\%, 140\%, 68\%, 157\%, 160\%, and 177\% respectively over the recall rates of the comparison method based on Shape Contexts.

\subsection{Importance of Free-Space}
\label{sec:results_distance}

\begin{figure}[t] 
    \centering
    \begin{subfigure}[b]{0.45\columnwidth}
    \includegraphics[trim={0cm 0cm 0cm 0cm},clip,width=\columnwidth]{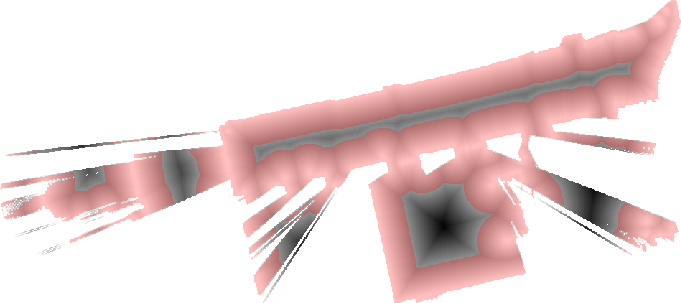}
        \caption{}
        \label{fig:mask_75}
    \end{subfigure}
    \begin{subfigure}[b]{0.45\columnwidth}
    \includegraphics[trim={0cm 0cm 0cm 0cm},clip,width=\columnwidth]{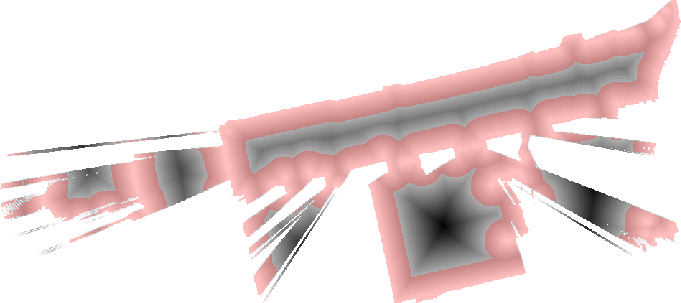}
        \caption{}
        \label{fig:mask_50}
    \end{subfigure}
    \begin{subfigure}[b]{0.45\columnwidth}
    \includegraphics[trim={0cm 0cm 0cm 0cm},clip,width=\columnwidth]{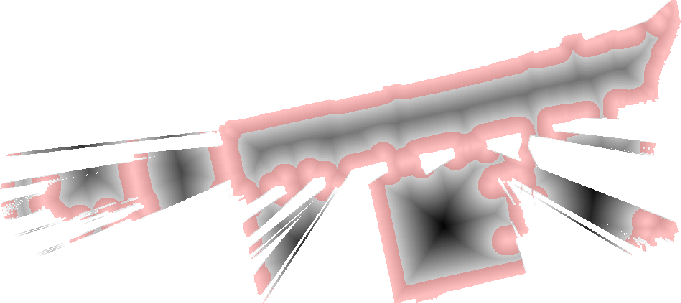}
        \caption{}
        \label{fig:mask_25}
    \end{subfigure}
    \begin{subfigure}[b]{0.45\columnwidth}
    \includegraphics[trim={0cm 0cm 0cm 0cm},clip,width=\columnwidth]{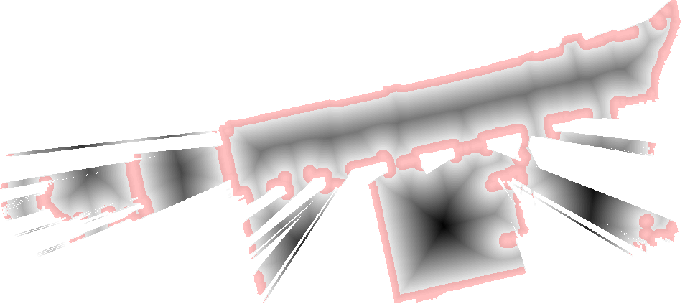}
        \caption{}
        \label{fig:mask_10}
    \end{subfigure}
    \caption{An example submap with with various permissible regions for feature extraction for the experiment described in Sec~\ref{sec:results_distance}. Submaps (\subref{fig:mask_75}), (\subref{fig:mask_50}), (\subref{fig:mask_25}), (\subref{fig:mask_10}) show settings of $d_{\text{threshold}}$ of 2.0\,m, 1.5\,m, 1.0\,m and 0.5\,m respectively.}
    \label{fig:masks}
    \vspace{-4mm}
\end{figure}

In this section we aim to determine the contribution of features in free-space to the performance of the proposed method. We perform a similar analysis to the one described in Sec.~\ref{sec:results_descriptor_performance}, however we limit the proposed method to extracting features near surface boundaries, to varying degrees. In particular, we perform several trails, removing features further than some distance $d_{\text{threshold}}$ from a surface boundary. We generate performance curves for settings of this threshold $d_{\text{threshold}} \in \{2.0, 1.5, 1.0, 0.5\}$ meters. Figure~\ref{fig:masks} shows an example submap and the areas where extracted features will be kept in the submap's description for each trial.

\begin{figure}[t] 
    \centering
    \includegraphics[trim={0cm 0cm 0cm 0cm},clip,width=0.8\columnwidth]{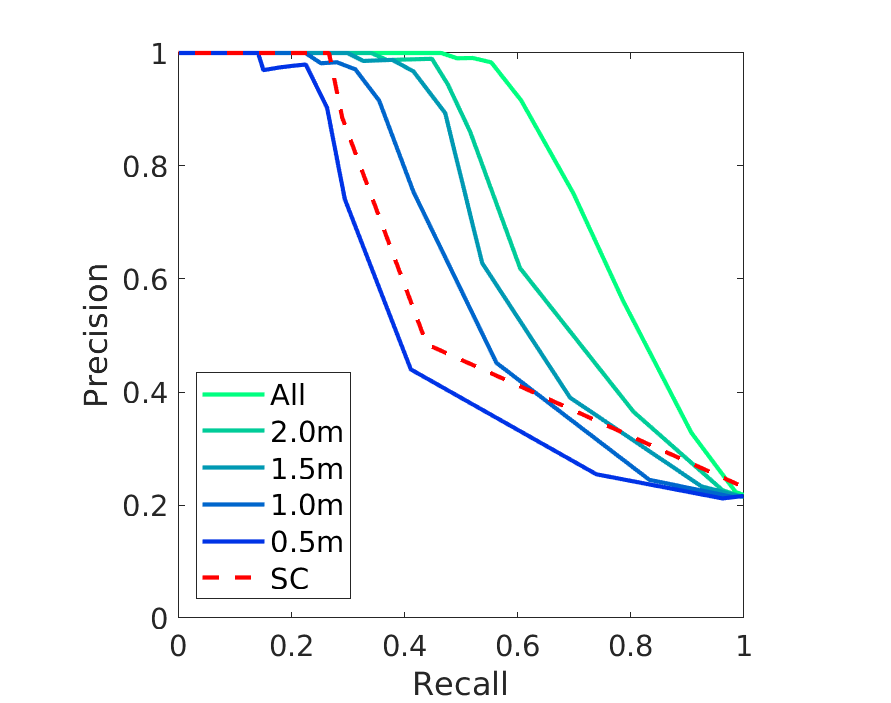}
    \caption{Precision Recall curves for a range of distances from surface boundaries in which frees-space features were extracted (see Sec.~\ref{sec:results_distance}. The plot indicates better matching performance as more free-space is included in the submap description. For comparison, the matching performance of Shape Contexts~\cite{belongie2002shape} is also included.}
    \label{fig:precision_recall_distances}
    \vspace{-3mm}
\end{figure}

Figure~\ref{fig:precision_recall_distances} shows the results of this analysis alongside the performance curve for the Shape Contexts method which is included for comparison. The results show that as we allow feature extraction at distances further from object surfaces the system's performance improves, as one might expect, indicating the utility of features located in free-space. The performance of Shape Contexts lies in the range of the performance of the proposed system when limited to a $\approx1$\,m region around surface boundaries. This agrees with intuition: keypoints extracted only on surfaces perform similarly to our method under the same limitation. Furthermore, this evaluation demonstrates that the performance of our proposal is due to the use of free-space, and not an advantage in the descriptive power of the keypoint.

\subsection{Place Recognition Experiment}

\begin{figure}[t] 
    \centering
    \includegraphics[trim={0cm 0cm 0cm 0cm},clip,width=\columnwidth]{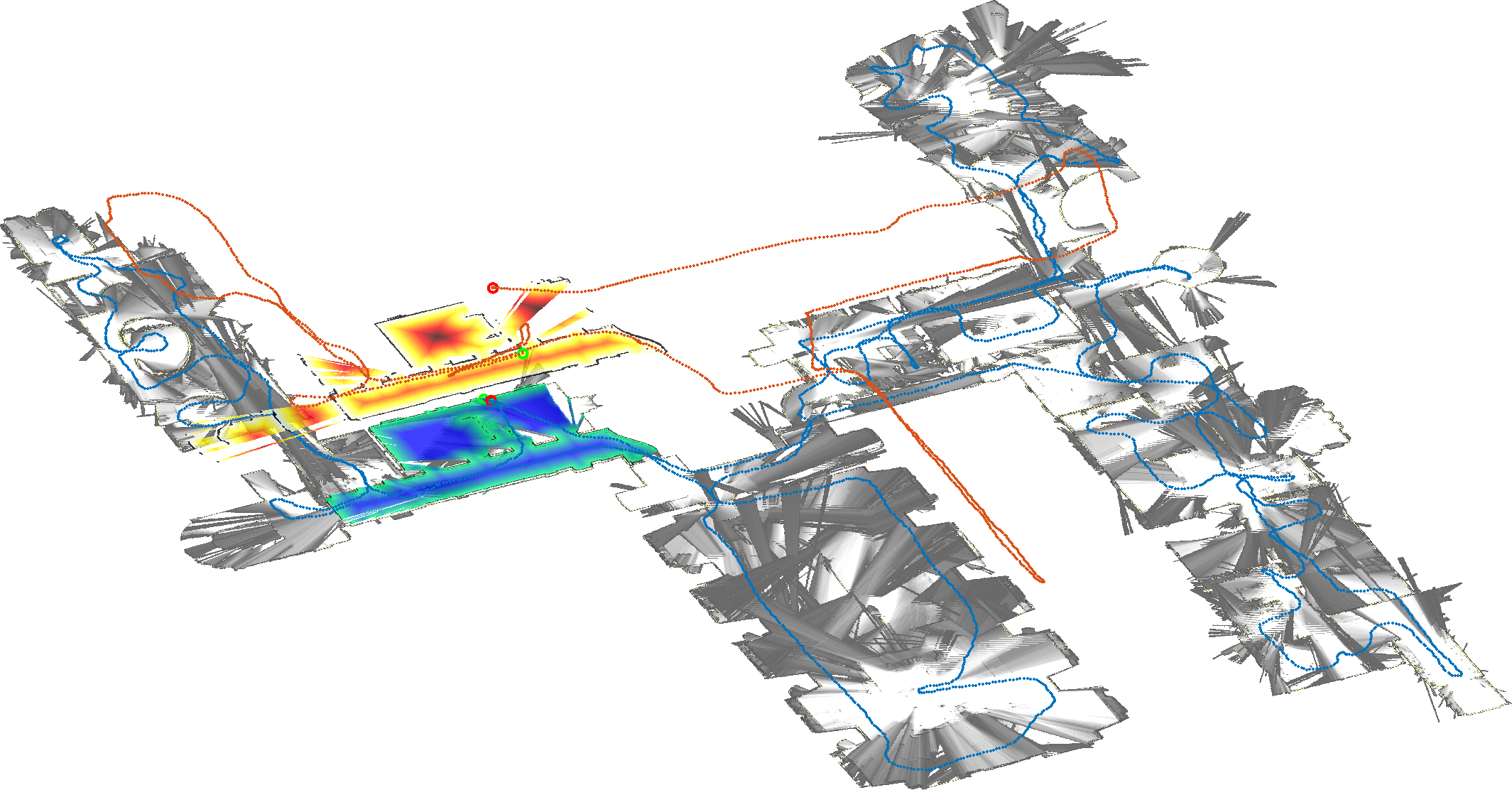}
    \caption{Localization experiment using the proposed method. Submaps created during traversal of the red trajectory are matched reference submaps from the blue trajectory. The experiment results in 292 submap-submap matches of which one is highlighted.}
    \label{fig:localization_experiment}
    \vspace{-5mm}
\end{figure}

We demonstrate the proposed approach in two place recognition experiments aimed at validating the proposed features in the context of a \ac{SLAM} problem. In the first experiment, we perform loop-closure detection, matching a submap against the collection of submaps generated earlier in the same trajectory. Figure~\ref{fig:teaser} shows one such positive match, between submaps at the beginning and end of the trajectory. In the second experiment, we perform localization of an agent in an existing map. For this purpose we used the data from two trajectories on the first-floor of the Deutsches Museum dataset\footnote{Database trajectory: b2-2016-04-27-12-31-41, Query Trajectory: b2-2016-04-05-14-44-52}. The experiment resulted in 292 submap-submap matches between a query map containing 53 submaps and a reference map containing 180 submaps. Note that each query submap can have multiple matches in the reference map. Figure~\ref{fig:localization_experiment} shows an example match with the query and reference trajectories aligned using the resulting transform. Of the 53 query submaps, 45 had at least one match to the reference map, and there were no false matches.

\section{Conclusion}
\label{sec:conclusion}

In this paper we proposed a novel approach for global localization in 2D lidar maps. At its core the system uses distance function representations of submaps, which allows extraction of novel features which describe the geometry of both occupied and non-occupied space. In particular, we use a \ac{DoH}-based detector to find points of high curvature on the \ac{SDF}. Keypoints are described using a gradient histogram, augmented with the feature distance, as well as the stationary point class. We test on publicly available datasets and demonstrate the efficacy on the proposed approach. Our tests show that the use of free-space improves localization performance when compared with using the proposed feature in the proximity of occupied space only. In addition, we compare against Shape Contexts, which showed state-of-the-art performance in a comparison of detector/descriptor combinations for submap characterization~\cite{bosse2009keypoint}. In our experiments, the proposed approach increases localization performance, measured as recall rate at precision 1.0, over the comparison method by an average of 92\% on the Deutsches Museum dataset and by 165\% on the PR2 office dataset. Future work will focus on determining if the results presented here are equally promising in 3D, where \ac{SDF} representations have seen wide application in recent years.

\bibliographystyle{ieeetr}
\bibliography{bibliography}

\begin{thebibliography}{10}

\bibitem{cadena2016past}
C.~Cadena, L.~Carlone, H.~Carrillo, Y.~Latif, D.~Scaramuzza, J.~Neira, I.~Reid,
  and J.~J. Leonard, ``Past, present, and future of simultaneous localization
  and mapping: Toward the robust-perception age,'' {\em IEEE Transactions on
  Robotics}, vol.~32, no.~6, pp.~1309--1332, 2016.

\bibitem{lau2013efficient}
B.~Lau, C.~Sprunk, and W.~Burgard, ``Efficient grid-based spatial
  representations for robot navigation in dynamic environments,'' {\em Robotics
  and Autonomous Systems}, vol.~61, no.~10, pp.~1116--1130, 2013.

\bibitem{izadi2011kinectfusion}
S.~Izadi, D.~Kim, O.~Hilliges, D.~Molyneaux, R.~Newcombe, P.~Kohli, J.~Shotton,
  S.~Hodges, D.~Freeman, A.~Davison, {\em et~al.}, ``Kinectfusion: real-time 3d
  reconstruction and interaction using a moving depth camera,'' in {\em
  Proceedings of the 24th annual ACM symposium on User interface software and
  technology}, pp.~559--568, ACM, 2011.

\bibitem{frisken2000adaptively}
S.~F. Frisken, R.~N. Perry, A.~P. Rockwood, and T.~R. Jones, ``Adaptively
  sampled distance fields: A general representation of shape for computer
  graphics,'' in {\em Proceedings of the 27th annual conference on Computer
  graphics and interactive techniques}, pp.~249--254, ACM Press/Addison-Wesley
  Publishing Co., 2000.

\bibitem{grisetti2007improved}
G.~Grisetti, C.~Stachniss, W.~Burgard, {\em et~al.}, ``Improved techniques for
  grid mapping with rao-blackwellized particle filters,'' {\em IEEE
  Transactions on Robotics}, 2007.

\bibitem{hess2016real}
W.~Hess, D.~Kohler, H.~Rapp, and D.~Andor, ``Real-time loop closure in 2d lidar
  slam,'' in {\em Robotics and Automation (ICRA), 2016 IEEE International
  Conference on}, pp.~1271--1278, IEEE, 2016.

\bibitem{olson2015m3rsm}
E.~Olson, ``M3rsm: Many-to-many multi-resolution scan matching,'' in {\em 2015
  IEEE International Conference on Robotics and Automation (ICRA)},
  pp.~5815--5821, IEEE, 2015.

\bibitem{gutmann1999incremental}
J.-S. Gutmann and K.~Konolige, ``Incremental mapping of large cyclic
  environments,'' in {\em Computational Intelligence in Robotics and
  Automation, 1999. CIRA'99. Proceedings. 1999 IEEE International Symposium
  on}, pp.~318--325, IEEE, 1999.

\bibitem{olson2009real}
E.~B. Olson, ``Real-time correlative scan matching,'' in {\em 2009 IEEE
  International Conference on Robotics and Automation}, pp.~4387--4393, IEEE,
  2009.

\bibitem{sivic2003video}
J.~Sivic and A.~Zisserman, ``Video google: A text retrieval approach to object
  matching in videos,'' in {\em null}, p.~1470, IEEE, 2003.

\bibitem{walthelm2002new}
A.~Walthelm, ``A new approach to global self-localization with laser range
  scans in unstructured environments,'' in {\em Intelligent Vehicle Symposium,
  2002. IEEE}, vol.~1, pp.~202--208, IEEE, 2002.

\bibitem{tipaldi2010flirt}
G.~D. Tipaldi and K.~O. Arras, ``Flirt-interest regions for 2d range data,'' in
  {\em Robotics and Automation (ICRA), 2010 IEEE International Conference on},
  pp.~3616--3622, IEEE, 2010.

\bibitem{tipaldi2013geometrical}
G.~D. Tipaldi, L.~Spinello, and W.~Burgard, ``Geometrical flirt phrases for
  large scale place recognition in 2d range data,'' in {\em 2013 IEEE
  International Conference on Robotics and Automation}, pp.~2693--2698, IEEE,
  2013.

\bibitem{granstrom2011learning}
K.~Granstr{\"o}m, T.~B. Sch{\"o}n, J.~I. Nieto, and F.~T. Ramos, ``Learning to
  close loops from range data,'' {\em The international journal of robotics
  research}, vol.~30, no.~14, pp.~1728--1754, 2011.

\bibitem{li2017deep}
J.~Li, H.~Zhan, B.~M. Chen, I.~Reid, and G.~H. Lee, ``Deep learning for 2d scan
  matching and loop closure,'' in {\em Intelligent Robots and Systems (IROS),
  2017 IEEE/RSJ International Conference on}, pp.~763--768, IEEE, 2017.

\bibitem{bosse2004simultaneous}
M.~Bosse, P.~Newman, J.~Leonard, and S.~Teller, ``Simultaneous localization and
  map building in large-scale cyclic environments using the atlas framework,''
  {\em The International Journal of Robotics Research}, vol.~23, no.~12,
  pp.~1113--1139, 2004.

\bibitem{millane2018c}
A.~Millane, Z.~Taylor, H.~Oleynikova, J.~Nieto, R.~Siegwart, and C.~Cadena,
  ``C-blox: A scalable and consistent tsdf-based dense mapping approach,'' in
  {\em 2018 IEEE/RSJ International Conference on Intelligent Robots and Systems
  (IROS)}, pp.~995--1002, IEEE, 2018.

\bibitem{bosse2009keypoint}
M.~Bosse and R.~Zlot, ``Keypoint design and evaluation for place recognition in
  2d lidar maps,'' {\em Robotics and Autonomous Systems}, vol.~57, no.~12,
  pp.~1211--1224, 2009.

\bibitem{oleynikova2017voxblox}
H.~Oleynikova, Z.~Taylor, M.~Fehr, R.~Siegwart, and J.~Nieto, ``Voxblox:
  Incremental 3d euclidean signed distance fields for on-board mav planning,''
  in {\em Intelligent Robots and Systems (IROS), 2017 IEEE/RSJ International
  Conference on}, pp.~1366--1373, IEEE, 2017.

\bibitem{fossel20152d}
J.-D. Fossel, K.~Tuyls, and J.~Sturm, ``2d-sdf-slam: A signed distance function
  based slam frontend for laser scanners,'' in {\em Intelligent Robots and
  Systems (IROS), 2015 IEEE/RSJ International Conference on}, pp.~1949--1955,
  IEEE, 2015.

\bibitem{koch2016multi}
P.~Koch, S.~May, M.~Schmidpeter, M.~K{\"u}hn, C.~Pfitzner, C.~Merkl, R.~Koch,
  M.~Fees, J.~Martin, D.~Ammon, {\em et~al.}, ``Multi-robot localization and
  mapping based on signed distance functions,'' {\em Journal of Intelligent \&
  Robotic Systems}, vol.~83, no.~3-4, pp.~409--428, 2016.

\bibitem{maurer2003linear}
C.~R. Maurer, R.~Qi, and V.~Raghavan, ``A linear time algorithm for computing
  exact euclidean distance transforms of binary images in arbitrary
  dimensions,'' {\em IEEE Transactions on Pattern Analysis and Machine
  Intelligence}, vol.~25, no.~2, pp.~265--270, 2003.

\bibitem{bay2006surf}
H.~Bay, T.~Tuytelaars, and L.~Van~Gool, ``Surf: Speeded up robust features,''
  in {\em European conference on computer vision}, pp.~404--417, Springer,
  2006.

\bibitem{dalal2005histograms}
N.~Dalal and B.~Triggs, ``Histograms of oriented gradients for human
  detection,'' in {\em Computer Vision and Pattern Recognition, 2005. CVPR
  2005. IEEE Computer Society Conference on}, vol.~1, pp.~886--893, IEEE, 2005.

\bibitem{lowe1999object}
D.~G. Lowe, ``Object recognition from local scale-invariant features,'' in {\em
  Computer vision, 1999. The proceedings of the seventh IEEE international
  conference on}, vol.~2, pp.~1150--1157, Ieee, 1999.

\bibitem{belongie2002shape}
S.~Belongie, J.~Malik, and J.~Puzicha, ``Shape matching and object recognition
  using shape contexts,'' tech. rep., California University, San Diego, La
  Jolla, Dept. of Computer Science and Engineering, 2002.

\bibitem{mason2012object}
J.~Mason and B.~Marthi, ``An object-based semantic world model for long-term
  change detection and semantic querying,'' in {\em 2012 IEEE/RSJ International
  Conference on Intelligent Robots and Systems}, pp.~3851--3858, IEEE, 2012.

\end{thebibliography}

\end{document}